\documentclass[times,twocolumn,final,authoryear]{elsarticle}

\usepackage{prletters}
\usepackage{framed,multirow}

\usepackage{amssymb}
\usepackage{latexsym}

\usepackage{url}
\usepackage{caption}
\usepackage{subcaption}

\usepackage{dirtytalk}

 \usepackage{multirow}

\usepackage{breakurl}
\usepackage[breaklinks]{hyperref}

\journal{Pattern Recognition Letters}

\begin{document}

\thispagestyle{empty}

\begin{table*}[!th]

\begin{minipage}{.9\textwidth}
\baselineskip12pt
\ifpreprint
  \vspace*{1pc}
\else
  \vspace*{-6pc}
\fi

\noindent {\LARGE\itshape Pattern Recognition Letters}
\vskip6pt

\noindent {\Large\bfseries Authorship Confirmation}

\vskip1pc

{\bf Please save a copy of this file, complete and upload as the
``Confirmation of Authorship'' file.}

\vskip1pc

As corresponding author
I, \underline{Fernando Alonso-Fernandez},
hereby confirm on behalf of all authors that:

\vskip1pc

\begin{enumerate}
\itemsep=3pt
\item This manuscript, or a large part of it, \underline {has not been
published,  was not, and is not being submitted to} any other journal.

\item If \underline {presented} at or \underline {submitted} to or
\underline  {published }at a conference(s), the conference(s) is (are)
identified and  substantial \underline {justification for
re-publication} is presented  below. A \underline {copy of
conference paper(s) }is(are) uploaded with the  manuscript.

\item If the manuscript appears as a preprint anywhere on the web, e.g.
arXiv,  etc., it is identified below. The \underline {preprint should
include a  statement that the paper is under consideration at Pattern
Recognition  Letters}.

\item All text and graphics, except for those marked with sources, are
\underline  {original works} of the authors, and all necessary
permissions for  publication were secured prior to submission of the
manuscript.

\item All authors each made a significant contribution to the research
reported  and have \underline {read} and \underline {approved} the
submitted  manuscript.
\end{enumerate}

Signature\underline{\hphantom{\hspace*{7cm}}} Date\underline{\hphantom{\hspace*{4cm}}}
\vskip1pc

\rule{\textwidth}{2pt}
\vskip1pc

{\bf List any pre-prints:}
\vskip5pc

\rule{\textwidth}{2pt}
\vskip1pc

{\bf Relevant Conference publication(s) (submitted, accepted, or
published):}
\vskip5pc

{\bf Justification for re-publication:}

\end{minipage}
\end{table*}

\clearpage
\thispagestyle{empty}
\ifpreprint
  \vspace*{-1pc}
\fi

\begin{table*}[!th]
\ifpreprint\else\vspace*{-5pc}\fi

\section*{Graphical Abstract (Optional)}
To create your abstract, please type over the instructions in the
template box below.  Fonts or abstract dimensions should not be changed
or altered.

\vskip1pc
\fbox{
\begin{tabular}{p{.4\textwidth}p{.5\textwidth}}
\bf On the Effect of Selfie Beautification Filters on Face Detection and Recognition  \\
Pontus Hedman, Vasilios Skepetzis, Kevin Hernandez-Diaz, Josef Bigun, Fernando Alonso-Fernandez \\[1pc]
\includegraphics[width=.3\textwidth]{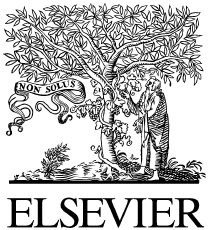}
&
This is the dummy text for graphical abstract.
%}\\
\end{tabular}
}

\end{table*}

\clearpage
\thispagestyle{empty}

\ifpreprint
  \vspace*{-1pc}
\else
%  \vspace*{-6pc}
\fi

\begin{table*}[!t]
\ifpreprint\else\vspace*{-15pc}\fi

\section*{Research Highlights (Required)}

To create your highlights, please type the highlights against each
\verb+\item+ command.

\vskip1pc

\fboxsep=6pt
\fbox{
\begin{minipage}{.95\textwidth}
It should be short collection of bullet points that convey the core
findings of the article. It should  include 3 to 5 bullet points
(maximum 85 characters, including spaces, per bullet point.)
\vskip1pc
\begin{itemize}

  \item We summarise works in image digital manipulation with the purpose of facial beautification

  \item We study the impact of enhancement and Augmented Reality filters on face detection and recognition

  \item We develop a method to reverse the applied manipulations that entail eye obfuscation

  \item We study if training the recognition system with manipulated images helps to increase accuracy

\end{itemize}
\vskip1pc
\end{minipage}
}

\end{table*}

\clearpage

\ifpreprint
  \setcounter{page}{1}
\else
  \setcounter{page}{1}
\fi

\begin{frontmatter}

\title{On the Effect of Selfie Beautification Filters on Face Detection and Recognition}

\author[1]{Pontus \snm{Hedman}}
\author[1]{Vasilios \snm{Skepetzis}}
\author[1]{Kevin \snm{Hernandez-Diaz}}
\author[1]{Josef \snm{Bigun}}
\author[1]{Fernando \snm{Alonso-Fernandez}\corref{cor1}}
\cortext[cor1]{Corresponding author:
  Tel.: +46-(0)35-167100;
  fax: +46-(0)35-120348;}
\ead{feralo@hh.se}

\address[1]{School of Information Science, Computer and Electrical Engineering, Halmstad University, Box 823, Halmstad SE 301-18,
Sweden}

%\received{1 May 2013}
%\finalform{10 May 2013}
%\accepted{13 May 2013}
%\availableonline{15 May 2013}
%\communicated{S. Sarkar}

\begin{abstract}

Beautification and augmented reality filters are very popular in applications that use selfie images. % captured with smartphones or personal devices.
However, they can distort or modify biometric features, severely affecting the capability of recognizing individuals' identity or even detecting the face.
Accordingly, we address the effect of such filters on the accuracy of automated face detection and recognition.
The social media image filters studied either modify the image contrast or illumination or occlude parts of the face. % with for example artificial glasses or animal noses.
We observe that the effect of some of these filters is harmful both to face detection and identity recognition, specially if they obfuscate the eye or (to a lesser extent) the nose.
To counteract such effect, we develop a method to reverse the applied manipulation with a modified version of the U-NET segmentation network. This is observed to contribute to a better face detection and recognition accuracy.
From a recognition perspective, we employ distance measures and trained machine learning algorithms applied to features extracted using several CNN backbones.  % a ResNet34 network. % trained to recognize faces.
We also evaluate if incorporating filtered images to the training set of machine learning approaches are beneficial. % for identity recognition.
Our results show good recognition when filters do not occlude important landmarks, specially the eyes. % (identification accuracy $>$99\%, EER$<$2\%).
The combined effect of the proposed approaches also allow to mitigate the impact produced by filters that occlude parts of the face.

\end{abstract}

\begin{keyword}
%\MSC 41A05\sep 41A10\sep 65D05\sep 65D17
\KWD Face Detection\sep Face Recognition\sep Social Media Filters\sep Beautification\sep U-NET\sep Convolutional Neural Network

%% MSC codes here, in the form: \MSC code \sep code
%% or \MSC[2008] code \sep code (2000 is the default)
\end{keyword}

\end{frontmatter}

%\linenumbers

\section{Introduction}

Selfie images captured with smartphones enjoy huge popularity and acceptability, and social media platforms centered around sharing such images offer several filters to ''beautify'' them before uploading.
%
%It has been shown that
Filtered images are more likely to be viewed and commented, achieving a higher engagement \citep{bakhshi_why_nodate}. %, and
Selfies are also increasingly used in security applications since mobiles have become data hubs used for all type of transactions \citep{[Rattani19selfiechIntroSelfie]}. Even
%
%The boom of online activity due to the pandemic has caused that
video conference applications, which have boomed during the pandemic, include beautification or augmented reality filters too.

A challenge posed by such filters is that facial features may be distorted or concealed.
Given their low cost and instant availability, they are a commodity used daily by many, not necessarily with the aim of compromising face recognition systems.
However, the capability of recognizing individuals may be affected, and even the possibility of detecting the face itself before any recognition can take place.
This is crucial for example in crime investigation on social media \citep{powell_social_2020}, where automatic pre-analysis is necessary given the magnitude of information posted or stored in confiscated devices \citep{hassan_gathering_2019}.
There are multiple examples of crimes captured on mobiles \citep{berman_hate_nodate,pagones_live-streamed_2021}, with the most striking lately being the use of posted videos of the US Capitol to identify rioters \citep{morrison_capitol_2021}.

There is, therefore, interest to study the consequences of different levels of image manipulation and concealment of facial parts due to these ''beautification'' filters. %,
%distortion %social media effects
%both on facial detection and recognition systems.
%
%The different levels refer to the degree of facial concealment or manipulation of image contrast and illumination, which could be critical to automated facial systems.
%
It would be also of interest to evaluate methods that remove the filter's effect to avoid a decrease in face detection and recognition performance.
The purpose and contributions of this work are therefore multi-fold.
We first summarize related works in image digital manipulation, in particular with the purpose of facial beautification.
Then, we study the impact of image enhancement and Augmented Reality (AR) filters both on the detection of filtered faces and on the recognition of individuals.
To counteract the effect of such filters, we develop a method to reverse some of the applied manipulations.
We focus on reversing modifications to the eye region, since these are observed to have the biggest impact on face detection or recognition.
Another strategy that we propose is the use of manipulated images to train the identity recognition system, which is also observed to increase accuracy when such manipulations are present in test images as well.

\begin{figure}[t]
    \centering
    \includegraphics[width=0.44\textwidth]{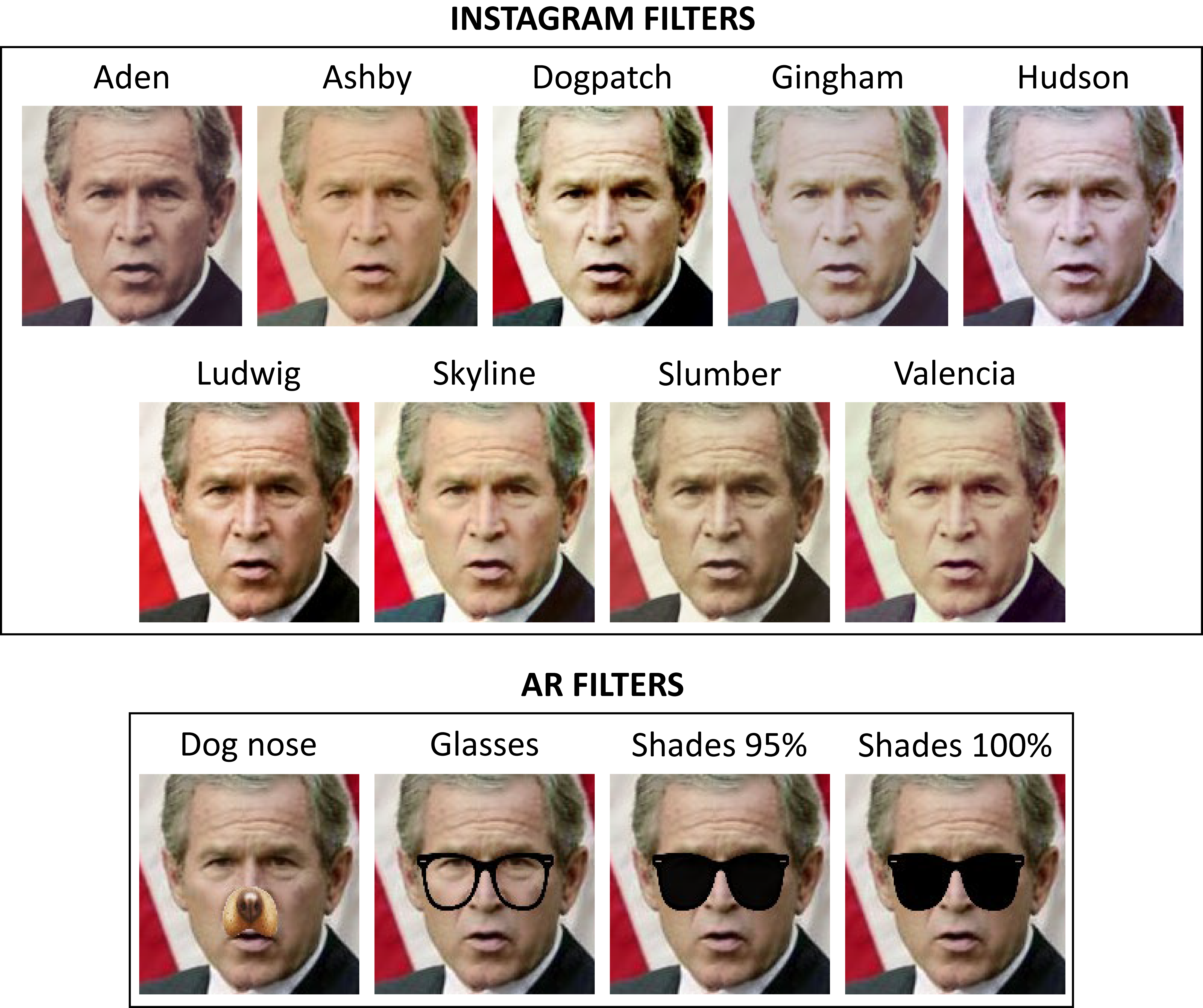}
\caption{Examples of the filters applied.}
\label{fig:filters}
\end{figure}

\section{Related Works}
\label{sec:soa}

%Studies focusing on facial manipulation either try to identify manipulated images, to measure the effect of %the manipulation, %on facial recognition,
%or to remove the manipulation.
%
%As shown in Figure~\ref{fig:soa},
%
Facial manipulation can be done in the physical or digital domain.
Physical manipulation can be achieved for example via make-up or surgery, while digital manipulation or retouching is via software \citep{rathgeb_impact_2019}.
Physical manipulation can be permanent (surgery) or non-permanent (make-up). Make-up can be quickly induced, so the same person may appear different even after a short period. Also, given the wide acceptance of cosmetics, it may also appear in enrolment data.
Digital retouching allows similar modifications than surgery or cosmetics, but in the digital domain, as well as other changes such as re-positioning or resizing facial landmarks.
A common aim of these modifications is to improve attractiveness (beautification).
Of course, it is also possible that someone
%
%
%Although, of course, the possibility exists that someone
pretends to look like somebody else to gain illegitimate access, or to hide the own identity to avoid recognition \citep{RamachandraPADsurvey,ScherhagMAsurvey}.
Another manipulation is the use of facial masks, either surgical due to the current pandemic \citep{DamerMasksBiosig}, or artificial as used in Presentation Attacks \citep{RamachandraPADsurvey}. However, this is out of the scope of this paper, since they are not oriented towards beautification.

Some works focus on detecting retouched images. %include
%\citep{bharati_detecting_2016,bharati_demography-based_2017,jain_detecting_2019,rathgeb_differential_2020}.
%
The methods proposed include
Supervised Restricted Boltzmann Machine (SRBM) \citep{bharati_detecting_2016},
semi-supervised autoencoders \citep{bharati_demography-based_2017},
or Convolutional Neural Networks (CNN) \citep{jain_detecting_2019}.
They also present new databases such as the ND-IIITD Retouched Faces Database \citep{bharati_detecting_2016} or the MDRF Multi-Demographic Retouched Faces \citep{bharati_demography-based_2017}, generated with paid and free applications that provide for example skin smoothing, face slimming, eye/lip color change, eye/teeth brightening, etc.
Authors of the MDRF database also analyze the impact of gender or ethnicity, showing that detection accuracy can vary greatly with the demographics.
The work \citep{jain_detecting_2019} also analyzes the detection of GAN-altered images.
All these approaches consider the use of a single image (the retouched one).
%
%On the other hand,
In contrast, \cite{rathgeb_differential_2020} proposes a differential approach where the unaltered image is also available, something which, according to the authors, is plausible in some scenarios (e.g. border control).
They use texture and deep features with images of the FERET and FRGCv2 datasets. Retouching is done with free applications from the Google PlayStore, arguing that free applications are more likely to be used by consumers.

Another set of works analyze the impact of manipulated images on the recognition performance. %\citep{dantcheva_can_2012,ferrara_impact_2013,bharati_detecting_2016,rathgeb_differential_2020}.
In \citep{dantcheva_can_2012}, they gather two databases of Caucasian females with makeup. One is from before/after YouTube tutorials mostly affecting the ocular area, %and with factors such as illumination, background or hair reasonably constant across shots of the same person.
and the other is by modification of FRGC images with lipstick, eye makeup or full makeup. The study employs Gabor features, LBP and a commercial system, showing an increase in error when testing against makeup pictures. They also found that applying LBP to Gabor filtered images (as opposed to the original image) partly compensates the effect.
In \citep{ferrara_impact_2013}, alterations such as barrel distortion or aspect ratio change are studied. They also simulate surgery digitally, such as injectables, wrinkle removal, lip augmentation, etc.
They employ the AR face database, with two commercial and a SIFT algorithm, concluding that the systems can overcome limited alterations, but they stumble on heavy manipulations.
Digital retouching is studied in \citep{bharati_detecting_2016} with the ND-IIITD database. They use a commercial system and OpenBR, an open source face engine, finding that the performance is considerably degraded when testing against retouched images.
%
%In \citep{rathgeb_impact_2019}, an overview is given, grounded in earlier studies on the effect of plastic surgery and cosmetics on face recognition, and drew analogies with the digital version of these alterations.
%
Image retouching is also examined by \cite{rathgeb_differential_2020} with a commercial system and the open-source ArcFace, showing its negative impact as well. % on the recognition performance.

\section{Materials and Methods}

\subsection{Beautification Filters}
\label{sect:filters}

\begin{figure}[t]
    \centering
    \begin{subfigure}{.09\textwidth}
      \centering
      \includegraphics[width=\linewidth]{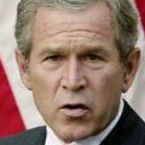}
      \caption{Original}
      \label{fig:original_face_image}
    \end{subfigure}
    \begin{subfigure}{.09\textwidth}
      \centering
      \includegraphics[width=\linewidth]{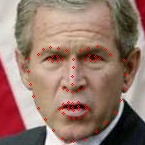}
      \caption{Landmarks}
      \label{fig:landmarks_face_image}
    \end{subfigure}
    \begin{subfigure}{.205\textwidth}
      \centering
      \includegraphics[width=\linewidth]{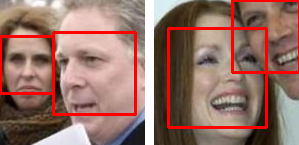}
      %\caption{Jean\_Charest\_0004}
      \caption{Multiple faces}
      \label{fig:fail_images_filter_apply}
    \end{subfigure}
    %
    %\begin{subfigure}{.09\textwidth}
    %  \centering
    %  \includegraphics[width=\linewidth]{Images/Marked_Faces_Julianne_Moore_0001.png}
    %  \caption{Julianne\_Moore\_0001}
    %\end{subfigure}
\caption{Example of: (a), (b) detected landmarks, (c) multiple faces detected.}
\label{fig:example_landmarks}
\end{figure}

We focus on two %types of %image forgery: enhancement and tampering.
manipulations: image enhancement and Augmented Reality (AR).
%
%These modifications, in particular
AR filters in particular have not been addressed in the literature.
For enhancement, we use the 9 most popular selfie Instagram filters \citep{noauthor_most_nodate}, which mostly change contrast and lighting (Figure~\ref{fig:filters}, top). The ranking is based on %the %number of
the images with each %a particular
filter and the hashtag ``\#selfie''. %As shown in Figure~\ref{fig:9_insta_filters}, they mostly change the contrast and lighting of the image.
Since the Instagram API does not allow to process %apply filters to
a large amount of data, the filters
%they
are recreated with a four-layer %fully connected
neural network %that learns the
%RGB, lightness, and saturation
%changes of each filter
\citep{hoppe_thoppeinstafilter_2021}.
%
%Since the API of Instagram does not allow to apply their filters to a large amounts of images, we have recreated them with the Instafilter library in Python \citep{hoppe_thoppeinstafilter_2021}. This library uses a four-layer fully connected neural network to learn the RGB, lightness, and saturation changes of various Instagram filters.  %\cite{hoppe_instagram_2020}.
%
Regarding %tampering filters, we make use of Augmented Reality (AR)
AR filters, they obfuscate face parts %of the face
that can be critical for recognition %(eye and nose,
(Figure~\ref{fig:filters}, bottom). Such filters are very popular in social media (e.g. Snapchat) and even in video conference platforms. % like Zoom. %In particular,
We apply: \say{Dog nose}, \say{Transparent glasses}, \say{Sunglasses-slight transparency}, and \say{Sunglasses-no transparency}. %For this purpose,
These %artificial nose or glasses
are merged with the face %image in the appropriate position
by using the landmarks (Figure~\ref{fig:landmarks_face_image}) given by
%the \say{face\_landmarks} function in the Python library \say{face\_recognition}
\citep{geitgey_face-recognition_nodate}.

\begin{figure}[t]
    \centering
    \includegraphics[width=0.48\textwidth]{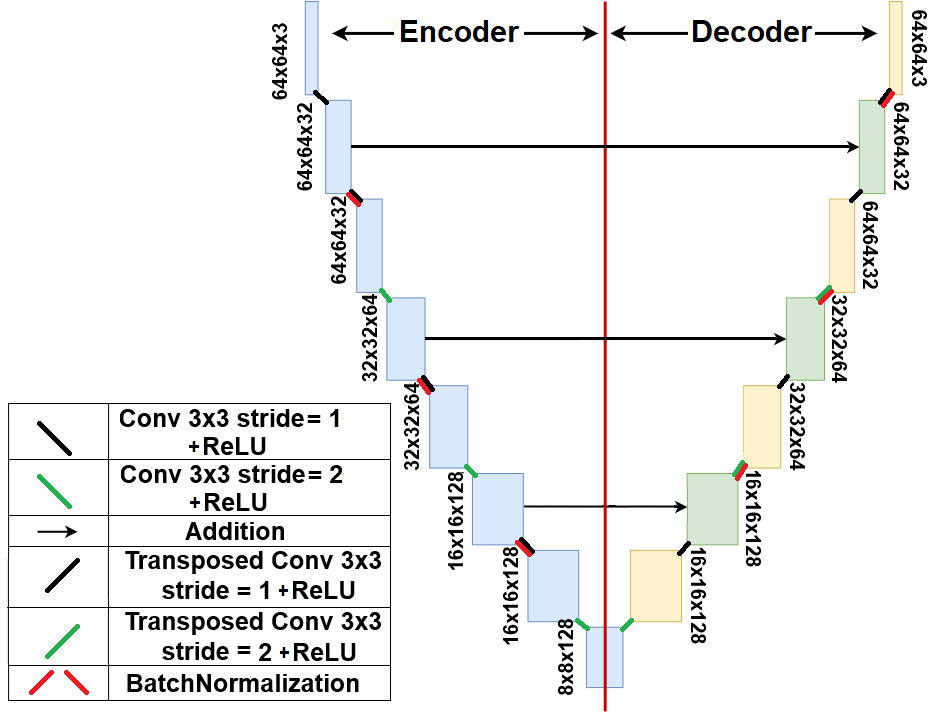}
    \caption{U-NET model employed. The blue rectangles are convolutions (compression), the yellow are transpose convolutions (expansion), and the green symbolizes the addition of blue and yellow through the add operation.}
    \label{fig:UNET}
\end{figure}

\subsection{Image Reconstruction with U-NET}
\label{sect:unet}

%To reconstruct the filtered images where the eyes have been obfuscated,
We use a modified version of the U-NET network.
%
%U-NET was
Originally presented for %biomedical
image segmentation \citep{ronneberger_u-net_2015},
it outperformed more complex networks in accuracy and speed while requiring less training data.
It has a compression or encoding path with convolutions and max-pooling, %consisting on a combination of convolutions, %with max-pooling layers,
followed by decompression or decoding path with up-convolutions. % to upsample feature maps. %consisting on transposed convolutions to up-sample the feature maps.
This gives the network a U-shape (Figure~\ref{fig:UNET}). %, hence its name.
Residual links connect maps of the encoding and decoding paths, %with the same size,
with channels concatenated, allowing the model to %ignore the parts of the image that remains the same and
focus on the parts of the image that change. % after applying the filters.
The original network has been modified, since the task is different. Inspired by \cite{springenberg_striving_2015}, max-pooling and up-convolutions are changed to strided convolutions/transposed convolutions. %, in an effort to stabilize the network training and performance.
Also, map concatenation in residual links is changed by addition to halve the number of channels. %
With this, we expect to still retain changes of image patches while counteracting over-fitting.

%\section{Experimental Framework}
%\label{sec:protocol}

%\subsection{Data Preparation and Processing}
\subsection{Databases}
%\section{Databases and Image Manipulation}
\label{sec:data}

%This section details the datasets employed and the method of creating the \textit{beautified} filtered images. It also presents the  U-NET network employed to regenerate images where the filters employed occlude parts of the face (in particular, the eyes).

%\subsection{Data}
%
%\subsection{Labeled Faces in the Wild (LFW) database}
%
We use the version aligned by funneling \citep{Huang2012a} %The main dataset is
of Labeled Faces in the Wild (LFW) \citep{LFWTech}. %\citep{LFWTech,LFWTechUpdate}.
%
%The dataset was created at the University of Massachusetts by Gary B. Huang, Manu Ramesh, Tamara Berg, and Erik Learned-Miller \cite{LFWTech}, and is made available for academic purposes\footnote{The dataset is made available here: \href{http://vis-www.cs.umass.edu/lfw/}{http://vis-www.cs.umass.edu/lfw/}}.
%
It has 13,233 images of 5,749 celebrities from the web with large variations in pose, light, expression, etc.
To ensure a sufficient amount of images per person, we remove people with less than 10 images, resulting in 158 individuals and 4,324 images.
%A crop of 145$\times$145 pixels around the center of each image is extracted, so background information around the faces is removed.
%
%We use the version with faces aligned by funneling \citep{Huang2012a}.
%
Five datasets are then created by applying the Instagram and AR filters of Section~\ref{sect:filters}.
The Instagram dataset is created by applying one filter %of Figure~\ref{fig:filters}
randomly to each unfiltered image.
Additionally, images with sunglasses are processed with the U-NET method of Section~\ref{sect:unet}, giving two more datasets of reconstructed images.
This results in 8 different datasets, %which are
listed in Table~\ref{tab:datasets}.
%
%Different beautification filters are then applied, as described in Section~\ref{sect:filters}, as well as the generative model of Section~\ref{sect:unet}.
%
%After that, biometric recognition experiments are done to evaluate the impact of the different filters.
%
%
%

%\subsection{CelebA database}
%
%
%The CelebA dataset ($202,599$ pictures of $10,177$ people) \citep{liu2015faceattributes} is used to train the model of Section~\ref{sect:unet}. % (we do not use LFW due to its limited amount of data).
%
U-NET is trained to reconstruct the filters shades\_leak and shades\_no\_leak (Figure~\ref{fig:filters}, bottom right) with the CelebA dataset ($202,599$ pictures of $10,177$ people) \citep{liu2015faceattributes}.
We use a batch size of 64, with Adam as optimizer and the \textit{MSE} between the output and the target (unfiltered) images as loss.
CelebA is not used for biometric recognition experiments, allowing to test the generalization ability of the U-NET model on unseen data.

\subsection{Face Detection and Feature Extraction Algorithms} % with ResNet}
\label{sect:feature_extraction}

The 8 datasets %created for this paper %of Table~\ref{tab:datasets}
are further encoded into a feature vector that will be used for biometric authentication.
First, faces are detected with the \say{face\_location} function of %the Python library \say{face\_recognition}
\citep{geitgey_face-recognition_nodate}, which is based on the dlib Python library.
The detector used is the more accurate \textit{CNN}, rather than the default \textit{HOG} model. %, which has less accuracy.
The CNN detector is trained with 7213 face images gathered from publicly available datasets including ImageNet, AFLW, Pascal VOC, VGG, WIDER, and face scrub. %\footnote{https://github.com/davisking/dlib-models}.
If more than one face or no face is found, the image is discarded (e.g. Figure~\ref{fig:fail_images_filter_apply}), and it is no further considered for feature extraction or recognition experiments.
%
%Note in Table~\ref{tab:datasets} that the accuracy of the detector varies across the datasets, suggesting that the applied manipulations have different impact (Section~\ref{sec:results}).
%
Our baseline feature extractor is a
%
%Feature extraction is done with a
ResNet34 model %\citep{[He16]}
of 29 convolutional layers with the %number of
filters per layer reduced by half \citep{king_high_nodate},
pre-trained from scratch for face recognition using $\sim$3M faces of 7485 identities from the VGG dataset, the face scrub dataset, and other web-scraped images. %
It uses as input images of 64$\times$64, and produces a 128-dimensional vector (taken from the next-to-last layer). % before the classification layer).
%
%For comparison purposes,
We also use two other available models, one based on the light SqueezeNet architecture (18 layers, 113$\times$113 input, 1000-dimensional vector) \citep{[Alonso20SqueezeFacePoseNet]} and a ResNet50 model (50 layers, 224$\times$224 input, 2048-dimensional vector) \citep{[Cao18vggface2]}, both pre-trained for face recognition using $\sim$8.41M faces of $>$100K identities from the very large VGGFace2 and MS-Celeb-1M datasets.
The latter two models are selected for comparison purposes with ResNet34 in order to assess the use of a light network suitable for mobile operation (SqueezeNet) and a much deeper ResNet50 network.

\subsection{Face Identification and Verification Protocol}

%We carry out both biometric identification and verification experiments.
%
For identification, we carry out both closed- and open-set experiments.
To find the closest subject of the database, we use both distance measures (Euclidean, Manhattan, and Cosine) and trained approaches (Support Vector Machines, SVMs \citep{cortes_support-vector_1995} and Extreme Gradient Boosting, XGBoost \citep{chen_xgboost_2016}).
Since SVM is a binary classifier, we adopt a \textit{one-vs-all} approach with multiple SVMs, taking the decision of the model that is the most confident.
XGBoost is multi-class, using softmax with cross-entropy loss as objective.
SVM is a widely employed classifier with good results in biometric authentication \citep{[Fierrez18]}, and XGBoost has wide adoption in the industry, % and a proven record efficiency and flexibility for real-world applications,
having obtained top rankings in recent machine learning challenges \citep{dmlc_2021}. Before the experiments, feature vectors are scaled with \textit{min-max} normalization, so each element is in the $[0,1]$ range.

To measure closed-set identification accuracy, we compute the False Negative Identification Rate (FNIR) \citep{Tabassi14openset}, which is the fraction of mated searches (i.e. where there is an
enrolled template for the search image) where the enrolled mate is not the closest subject of the database. % outside the top rank.
As accuracy metric, we report the Genuine Accept Rate (GAR), computed as GAR=1-FNIR, which measures the fraction of mated searches where the enrolled template is in the top rank. % and the score exceeds the threshold).
Open-set identification accuracy is quantified by reporting the False Positive Identification rate (FPIR) and the False Negative Identification Rate (FNIR) \citep{Tabassi14openset}.
FPIR (also called False Accept Rate, FAR) is the fraction of non-mated searches (i.e. there is no enrolled template for the search image) where one or more enrolled identities are returned at or above an specified score threshold. % $T$.
To compute FNIR in open-set, one must consider if the mated search is not in the top rank or if its comparison score is below the threshold.
%
%FNIR is the fraction of mated searches (i.e. where there is an enrolled template for the search image) where the enrolled mate is outside the top rank, or the comparison score is below the threshold. % $T$.
%
FPIR and FNIR in open-set are obtained at different thresholds, after which we report the Detection Error Trade-off (DET), showing FPIR (FAR) against GAR. %FNIR.

%\subsection{Face Verification}

For verification experiments, we use the same distance measures than before (Euclidean, Manhattan, Cosine). Accuracy is measured via False Rejection Rate (FRR) and False Acceptance Rate (FAR) at different distance thresholds \citep{wayman_biometric_2009}.
Then, the %Detection Error Trade-off
DET curve is given, %which
plotting  %illustrate the trade-off between
FAR against FRR.
As a single measure of accuracy, we also report the Equal Error Rate (EER), which is the error at the threshold where FAR=FRR.

\begin{table}[t]
\small
    \caption{Summary description of the 8 datasets employed in the study. The numbers in brackets indicate the percentage of images of each dataset for which the face is correctly detected (see Section~\ref{sect:feature_extraction}).}
    \centering
    \begin{tabular}{ c|c|c }
    \textbf{Name} & \textbf{Manipulation} & \textbf{Images} \\
    \hline \hline
    benchmark & Original images & 4,276 (98.9\%) \\  \hline
    dog & Dog nose & 4,229 (97.8\%) \\
    glasses & Transparent glasses & 3,666 (84.8\%) \\
    instagram & Instagram filters & 4,277 (98.9\%) \\  \hline
    shades\_leak & Shades (95\% opacity) & 3,851 (89.1\%) \\
    shades\_recon\_leak & Reconstructed (95\% op) &  4,288 (99.2\%) \\ \hline
    shades\_no\_leak & Shades (100\% opacity) &  3,825 (88.5\%) \\
    shades\_recon\_no\_leak & Reconstructed (100\% op) &  4,271 (98.8\%)
    \end{tabular}
    \label{tab:datasets}
\end{table}

\begin{figure}[t]
    \centering
    \includegraphics[width=0.45\textwidth]{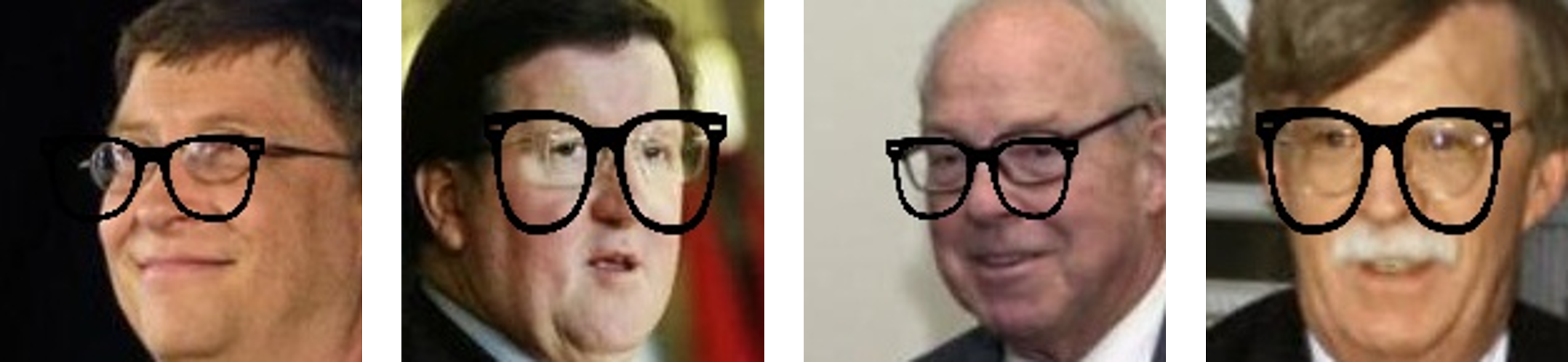}
    \caption{Example of images with AR glasses where the individual wears real glasses as well. In all these examples, the face detector fails.}
    \label{fig:detection-failures}
\end{figure}

\begin{figure}[t]
    \centering
    %\begin{subfigure}{.5\textwidth}
    %  \centering
    %  \includegraphics[width=.6\linewidth]{Images/original_censored.png}
    %  \caption{Original face image.}
    %  \label{fig:originalFaceImage}
    %\end{subfigure}
    %
    %\begin{subfigure}{.5\textwidth}
    %    \centering
    %    \includegraphics[width=.6\linewidth]{Images/shades_leak_censored.png}
    %    \caption{Shades filter applied to original image.}
    %    \label{fig:AppliedShades}
    %\end{subfigure}
    %%
    \begin{subfigure}{.2\textwidth}
      \centering
      \includegraphics[width=\linewidth]{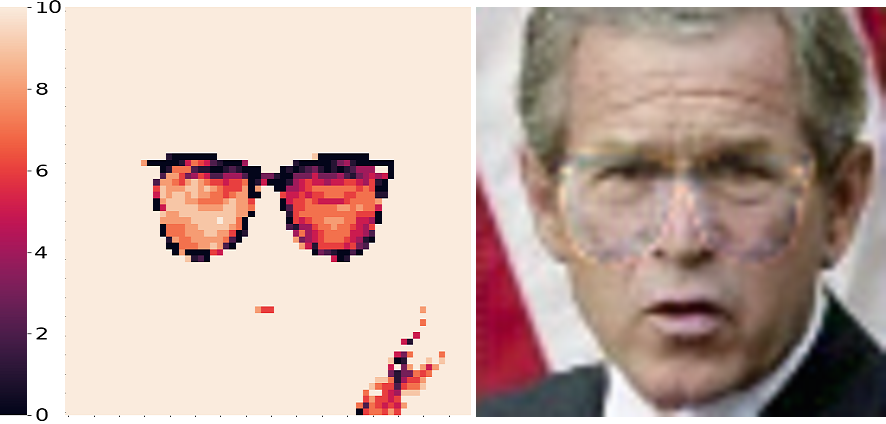}
      \caption{shades 95\%}
      \label{fig:sunglasses_reconstruct95}
    \end{subfigure}
    \begin{subfigure}{.2\textwidth}
      \centering
      \includegraphics[width=\linewidth]{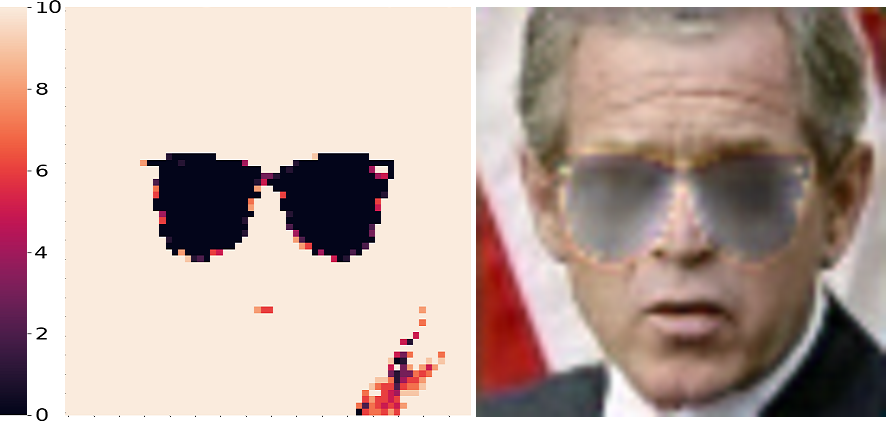}
      \caption{shades 100\%}
      \label{fig:sunglasses_reconstruct100}
    \end{subfigure}
\caption{Examples of the reconstruction on the shades dataset. The left part of each sub-figure shows the first 10 pixel values of the images with shades (re-scaled to 0-255), while the right part shows the reconstructed image. Shades at 95\% preserve some information, so a good reconstruction is still possible. Shades at 100\%, on the other hand, destroys the pixels behind the shades.}
\label{fig:example_reconstruction}
\end{figure}

\begin{figure*}[htb]
%\centering
%    \begin{subfigure}{.28\textwidth} %.16 for one line of figures only
%      \centering
%      \includegraphics[width=\linewidth]{Images/tsne_benchmark.png}
%      \caption{Benchmark}
%      \label{fig:benchmark}
%    \end{subfigure}
%    %
%    \begin{subfigure}{.28\textwidth}
%      \centering
%      \includegraphics[width=\linewidth]{Images/tsne_instagram.png}
%      \caption{Instagram}
%      \label{fig:instagram}
%    \end{subfigure}
%    %
%    \begin{subfigure}{.28\textwidth}
%      \centering
%      \includegraphics[width=\linewidth]{Images/tsne_dog.png}
%      \caption{Dog}
%      \label{fig:dog_nose}
%    \end{subfigure}
%    %
%    \begin{subfigure}{.28\textwidth}
%      \centering
%      \includegraphics[width=\linewidth]{Images/tsne_glasses.png}
%      \caption{Glasses}
%      \label{fig:tsne-glasses}
%    \end{subfigure}
%    %
%    \begin{subfigure}{.28\textwidth}
%      \centering
%      \includegraphics[width=\linewidth]{Images/tsne_shades_leak.png}
%      \caption{Shades 95\% Opacity}
%      \label{fig:sunglasses95}
%    \end{subfigure}
%    %
%    \begin{subfigure}{.28\textwidth}
%      \centering
%      \includegraphics[width=\linewidth]{Images/tsne_shades_no_leak.png}
%      \caption{Shades 100\% Opacity}
%      \label{fig:Sunglasses100}
%    \end{subfigure}
    %
    \centering
    \includegraphics[width=0.75\textwidth]{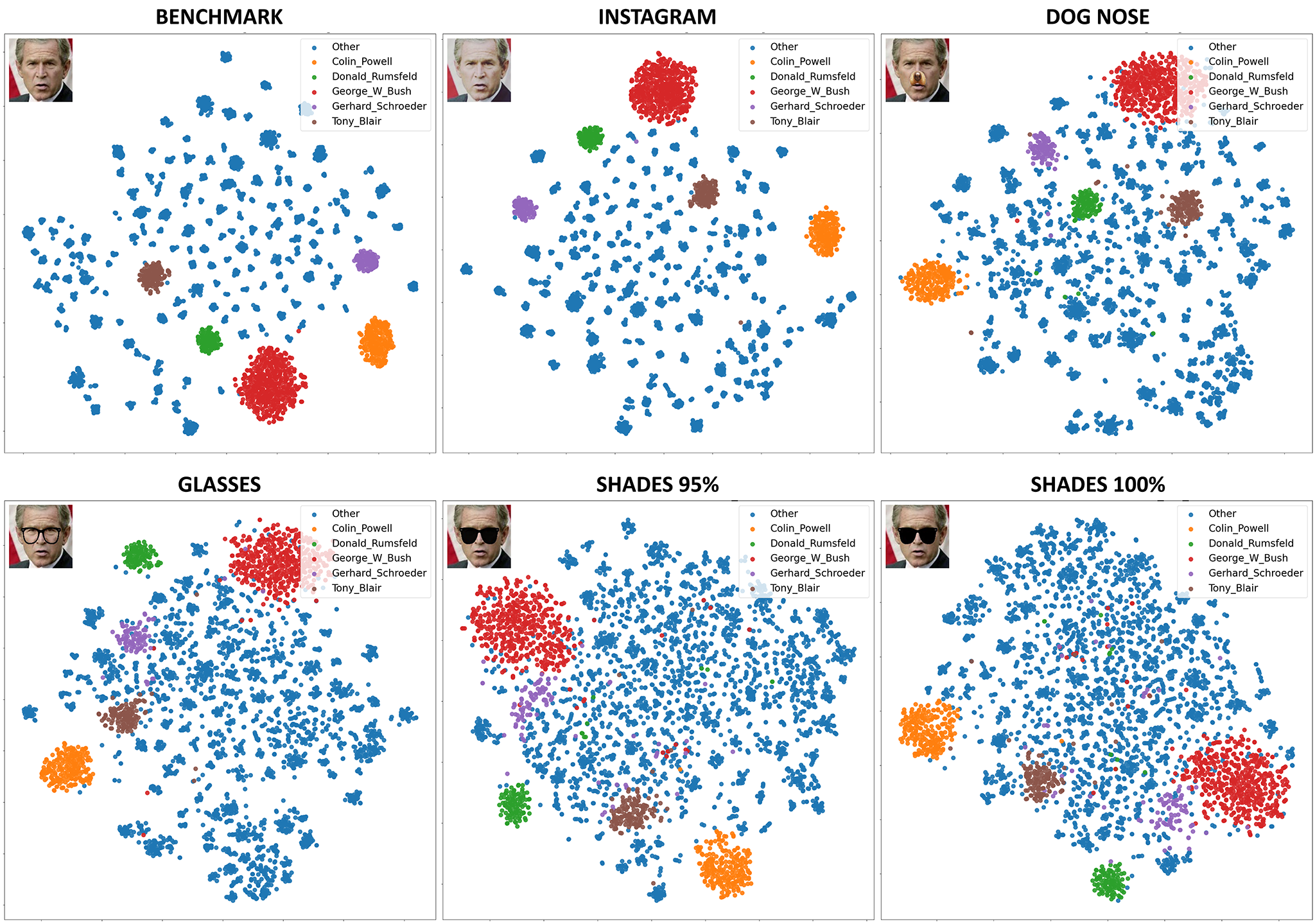}
\caption{Class separation for the various datasets shown via t-SNE scatter plots after feature extraction with ResNet34 (perplexity=30).}
\label{fig:t-sne_clusters}
\end{figure*}

\begin{table*}[t]
\footnotesize
    \caption{ResNet34: Identification Accuracy (closed set). The higher the value, the better.}
    \centering
    \begin{tabular}{ c||c|c|c||c|c|c||c|c|c|| }

    \multicolumn{1}{c||}{} & \multicolumn{3}{c||}{\textbf{Distance measures}} & \multicolumn{3}{c||}{\textbf{SVM}} & \multicolumn{3}{c||}{\textbf{XGBoost}} \\ \cline{2-10}

     &  &   &  &  \textbf{Train=} & \textbf{Train=} & \textbf{Train=} & \textbf{Train=} & \textbf{Train=} & \textbf{Train=} \\

    \textbf{Test Dataset} & \textbf{Eucl} & \textbf{Manh} & \textbf{Cosine}
    %& \textbf{Mean}
    & \textbf{Bench.} & \textbf{Filter} &  \textbf{All} & \textbf{Bench.} & \textbf{Filter} &  \textbf{All} \\

    \hline \hline

    benchmark                   &   \textbf{0.930}   &  0.921 & 0.925 %& 0.925
    & 0.993   &   N/A       &   \textbf{0.999} (+0.006)  & 0.993   &   N/A       &   \textbf{1.000} (+0.007) \\ \hline

    dog                         &   \textbf{0.562}   & 0.543 &  0.539 %& 0.548
    & 0.921   &   0.966 (+0.045)   &  \textbf{0.982} (+0.061)  & 0.917   &   0.974 (+0.057)   & \textbf{0.986} (+0.069) \\

    glasses                     &   0.479   & 0.455 &  \textbf{0.498} %& 0.477
    & 0.879   &   0.922 (+0.043)   &   \textbf{0.967} (+0.088)  &  0.892   &   0.925 (+0.033)   &   \textbf{0.967} (+0.075)  \\

    instagram                   &   \textbf{0.923}   & 0.916 &  0.921 %& 0.92
    & 0.992   &   0.991 (-0.001)   &   \textbf{0.998} (+0.006)  &  0.993   &   0.993 (+0.000)   &  \textbf{1.000} (+0.007) \\ \hline

    shades\_leak                &   \textbf{0.435}   & 0.407 &  0.423  %& 0.421
    & 0.708   &   0.866 (+0.158)   &  \textbf{0.964} (+0.256)  & 0.722   &   0.030 (-0.692)   &   \textbf{0.964} (+0.242) \\

    shades\_recon\_leak         &   \textbf{0.663}   & 0.639 &  0.655 %& 0.652
    & 0.885   &   0.946 (+0.061)   &   \textbf{0.949} (+0.064) &  0.881   &   0.932 (+0.051)   &    \textbf{0.937} (+0.056) \\ \hline

    shades\_no\_leak            &   \textbf{0.386}   & 0.365 &  0.379 %& 0.376
    & 0.672   &   0.854 (+0.182)   &   \textbf{0.941} (+0.269) & 0.663   &   0.009 (-0.654)   &   \textbf{0.948} (+0.285) \\

    shades\_recon\_no\_leak     &   \textbf{0.368}   & 0.357 & 0.350 %& 0.358
    & 0.594   &   \textbf{0.849} (+0.255)   &   0.827 (+0.233)  & 0.619   &   0.124 (-0.495)   &   \textbf{0.825} (+0.206)  \\ \hline \hline

     \textit{average} & \textbf{0.593}  & 0.575 & 0.586 & 0.831   &  0.913 (+0,082)  &  \textbf{0.953} (+0,122)  & 0.835  &   0.570 (-0,265)  &  \textbf{0.953} (+0,118)

    \end{tabular}
    \label{tab:identification}
\end{table*}

% Please add the following required packages to your document preamble:
% \usepackage{multirow}
\begin{table*}[]
\scriptsize
\caption{ResNet34: Identification Accuracy (closed set), cross-filter experiments. The higher the value, the better. A graphical representation is given in Figure~\ref{fig:identification-crossfilter}.}
\label{tab:identification-crossfilter}
\begin{tabular}{c||cccccccc||cccccccc||}
                                                              & \multicolumn{8}{c||}{\textbf{SVM}}                         & \multicolumn{8}{c||}{\textbf{XGBoost}}  \\ \cline{2-17}
\multicolumn{1}{c||}{}                                         & \multicolumn{8}{c||}{\textbf{Test Dataset}}  & \multicolumn{8}{c||}{\textbf{Test Dataset}}    \\ \hline \hline

\multicolumn{1}{|c||}{\multirow{2}{*}{\textbf{Train Dataset}}} & \multicolumn{1}{c|}{\multirow{2}{*}{\begin{tabular}[c]{@{}c@{}}bench-\\ mark\end{tabular}}} & \multicolumn{1}{c|}{\multirow{2}{*}{dog}} & \multicolumn{1}{c|}{\multirow{2}{*}{\begin{tabular}[c]{@{}c@{}}gla-\\ sses\end{tabular}}} & \multicolumn{1}{c|}{\multirow{2}{*}{\begin{tabular}[c]{@{}c@{}}insta-\\ gram\end{tabular}}} & \multicolumn{4}{c||}{shades-}                                                                                                                                                                                                                                                  & \multicolumn{1}{c|}{\multirow{2}{*}{\begin{tabular}[c]{@{}c@{}}bench-\\ mark\end{tabular}}} & \multicolumn{1}{c|}{\multirow{2}{*}{dog}} & \multicolumn{1}{c|}{\multirow{2}{*}{\begin{tabular}[c]{@{}c@{}}gla-\\ sses\end{tabular}}} & \multicolumn{1}{c|}{\multirow{2}{*}{\begin{tabular}[c]{@{}c@{}}insta-\\ gram\end{tabular}}} & \multicolumn{4}{c||}{shades-}                                                                                                                                                                                                                                                 \\ \cline{6-9} \cline{14-17}
\multicolumn{1}{|c||}{}                                        & \multicolumn{1}{c|}{}                                                                       & \multicolumn{1}{c|}{}                     & \multicolumn{1}{c|}{}                                                                     & \multicolumn{1}{c|}{}                                                                       & \multicolumn{1}{c|}{leak}           & \multicolumn{1}{c|}{\begin{tabular}[c]{@{}c@{}}recon-\\ leak\end{tabular}} & \multicolumn{1}{c|}{\begin{tabular}[c]{@{}c@{}}no-\\ leak\end{tabular}} & \multicolumn{1}{c||}{\begin{tabular}[c]{@{}c@{}}recon-\\ no-\\ leak\end{tabular}} & \multicolumn{1}{c|}{}                                                                       & \multicolumn{1}{c|}{}                     & \multicolumn{1}{c|}{}                                                                     & \multicolumn{1}{c|}{}                                                                       & \multicolumn{1}{c|}{leak}          & \multicolumn{1}{c|}{\begin{tabular}[c]{@{}c@{}}recon-\\ leak\end{tabular}} & \multicolumn{1}{c|}{\begin{tabular}[c]{@{}c@{}}no-\\ leak\end{tabular}} & \multicolumn{1}{c||}{\begin{tabular}[c]{@{}c@{}}recon-\\ no-\\ leak\end{tabular}} \\ \hline \hline

\multicolumn{1}{|c||}{benchmark}                               & \multicolumn{1}{c|}{\textbf{0.993}}                                                         & \multicolumn{1}{c|}{0.921}                & \multicolumn{1}{c|}{0.879}                                                                & \multicolumn{1}{c|}{0.992}                                                                  & \multicolumn{1}{c|}{0.708}          & \multicolumn{1}{c|}{0.885}                                                 & \multicolumn{1}{c|}{0.672}                                              & \multicolumn{1}{c||}{0.594}                                                       & \multicolumn{1}{c|}{\textbf{0.993}}                                                         & \multicolumn{1}{c|}{0.917}                & \multicolumn{1}{c|}{0.892}                                                                & \multicolumn{1}{c|}{0.993}                                                                  & \multicolumn{1}{c|}{0.722}         & \multicolumn{1}{c|}{0.881}                                                 & \multicolumn{1}{c|}{0.663}                                              & \multicolumn{1}{c||}{0.619}                                                       \\ \hline
\multicolumn{1}{|c||}{dog}                                     & \multicolumn{1}{c|}{0.973}                                                                  & \multicolumn{1}{c|}{\textbf{0.966}}       & \multicolumn{1}{c|}{0.762}                                                                & \multicolumn{1}{c|}{0.979}                                                                  & \multicolumn{1}{c|}{0.523}          & \multicolumn{1}{c|}{0.798}                                                 & \multicolumn{1}{c|}{0.472}                                              & \multicolumn{1}{c||}{0.482}                                                       & \multicolumn{1}{c|}{0.973}                                                                  & \multicolumn{1}{c|}{\textbf{0.974}}       & \multicolumn{1}{c|}{0.76}                                                                 & \multicolumn{1}{c|}{0.975}                                                                  & \multicolumn{1}{c|}{0.498}         & \multicolumn{1}{c|}{0.791}                                                 & \multicolumn{1}{c|}{0.444}                                              & \multicolumn{1}{c||}{0.463}                                                       \\ \hline
\multicolumn{1}{|c||}{glasses}                                 & \multicolumn{1}{c|}{0.978}                                                                  & \multicolumn{1}{c|}{0.775}                & \multicolumn{1}{c|}{\textbf{0.922}}                                                       & \multicolumn{1}{c|}{0.97}                                                                   & \multicolumn{1}{c|}{0.837}          & \multicolumn{1}{c|}{0.929}                                                 & \multicolumn{1}{c|}{0.775}                                              & \multicolumn{1}{c||}{0.745}                                                       & \multicolumn{1}{c|}{0.974}                                                                  & \multicolumn{1}{c|}{0.764}                & \multicolumn{1}{c|}{\textbf{0.925}}                                                       & \multicolumn{1}{c|}{0.972}                                                                  & \multicolumn{1}{c|}{0.813}         & \multicolumn{1}{c|}{0.91}                                                  & \multicolumn{1}{c|}{0.783}                                              & \multicolumn{1}{c||}{0.719}                                                       \\ \hline
\multicolumn{1}{|c||}{instagram}                               & \multicolumn{1}{c|}{0.993}                                                                  & \multicolumn{1}{c|}{0.918}                & \multicolumn{1}{c|}{0.888}                                                                & \multicolumn{1}{c|}{\textbf{0.991}}                                                         & \multicolumn{1}{c|}{0.716}          & \multicolumn{1}{c|}{0.879}                                                 & \multicolumn{1}{c|}{0.661}                                              & \multicolumn{1}{c||}{0.613}                                                       & \multicolumn{1}{c|}{0.994}                                                                  & \multicolumn{1}{c|}{0.889}                & \multicolumn{1}{c|}{0.871}                                                                & \multicolumn{1}{c|}{\textbf{0.993}}                                                         & \multicolumn{1}{c|}{0.699}         & \multicolumn{1}{c|}{0.88}                                                  & \multicolumn{1}{c|}{0.655}                                              & \multicolumn{1}{c||}{0.618}                                                       \\ \hline
\multicolumn{1}{|c||}{shades\_leak}                             & \multicolumn{1}{c|}{0.893}                                                                  & \multicolumn{1}{c|}{0.564}                & \multicolumn{1}{c|}{0.896}                                                                & \multicolumn{1}{c|}{0.877}                                                                  & \multicolumn{1}{c|}{\textbf{0.866}} & \multicolumn{1}{c|}{0.831}                                                 & \multicolumn{1}{c|}{0.944}                                              & \multicolumn{1}{c||}{0.802}                                                       & \multicolumn{1}{c|}{0.008}                                                                  & \multicolumn{1}{c|}{0.008}                & \multicolumn{1}{c|}{0.01}                                                                 & \multicolumn{1}{c|}{0.008}                                                                  & \multicolumn{1}{c|}{\textbf{0.03}} & \multicolumn{1}{c|}{0.008}                                                 & \multicolumn{1}{c|}{0.008}                                              & \multicolumn{1}{c||}{0.008}                                                       \\ \hline
\multicolumn{1}{|c||}{*\_recon\_leak}                             & \multicolumn{1}{c|}{0.977}                                                                  & \multicolumn{1}{c|}{0.804}                & \multicolumn{1}{c|}{0.916}                                                                & \multicolumn{1}{c|}{0.97}                                                                   & \multicolumn{1}{c|}{0.772}          & \multicolumn{1}{c|}{\textbf{0.946}}                                        & \multicolumn{1}{c|}{0.73}                                               & \multicolumn{1}{c||}{0.767}                                                       & \multicolumn{1}{c|}{0.967}                                                                  & \multicolumn{1}{c|}{0.787}                & \multicolumn{1}{c|}{0.907}                                                                & \multicolumn{1}{c|}{0.961}                                                                  & \multicolumn{1}{c|}{0.77}          & \multicolumn{1}{c|}{\textbf{0.932}}                                        & \multicolumn{1}{c|}{0.699}                                              & \multicolumn{1}{c||}{0.758}                                                       \\ \hline
\multicolumn{1}{|c||}{*\_no\_leak}                                & \multicolumn{1}{c|}{0.863}                                                                  & \multicolumn{1}{c|}{0.512}                & \multicolumn{1}{c|}{0.858}                                                                & \multicolumn{1}{c|}{0.855}                                                                  & \multicolumn{1}{c|}{0.957}          & \multicolumn{1}{c|}{0.802}                                                 & \multicolumn{1}{c|}{\textbf{0.854}}                                     & \multicolumn{1}{c||}{0.805}                                                       & \multicolumn{1}{c|}{0.008}                                                                  & \multicolumn{1}{c|}{0.008}                & \multicolumn{1}{c|}{0.01}                                                                 & \multicolumn{1}{c|}{0.008}                                                                  & \multicolumn{1}{c|}{0.009}         & \multicolumn{1}{c|}{0.008}                                                 & \multicolumn{1}{c|}{\textbf{0.009}}                                     & \multicolumn{1}{c||}{0.008}                                                       \\ \hline
\multicolumn{1}{|c||}{*\_recon\_no\_leak}                          & \multicolumn{1}{c|}{0.838}                                                                  & \multicolumn{1}{c|}{0.53}                 & \multicolumn{1}{c|}{0.79}                                                                 & \multicolumn{1}{c|}{0.82}                                                                   & \multicolumn{1}{c|}{0.839}          & \multicolumn{1}{c|}{0.857}                                                 & \multicolumn{1}{c|}{0.817}                                              & \multicolumn{1}{c||}{\textbf{0.849}}                                              & \multicolumn{1}{c|}{0.006}                                                                  & \multicolumn{1}{c|}{0.006}                & \multicolumn{1}{c|}{0.005}                                                                & \multicolumn{1}{c|}{0.006}                                                                  & \multicolumn{1}{c|}{0.005}         & \multicolumn{1}{c|}{0.006}                                                 & \multicolumn{1}{c|}{0.004}                                              & \multicolumn{1}{c||}{\textbf{0.124}}                                              \\ \hline
\end{tabular}

\end{table*}

\section{Results and Discussion}
\label{sec:results}

Table~\ref{tab:datasets} provides the number of images of each dataset for which a face is detected. %, so a feature vector is extracted.
Note that the detection accuracy varies across datasets, suggesting that the applied manipulations have different impact.
%
%It can be seen that
The benchmark (unfiltered) dataset has a detection rate of $\sim$99\%. The face is also detected successfully in case of Instagram filters, which can be expected since they mainly enhance contrast or lighting. %Interestingly,
Faces with a dog nose are also well detected, but occlusions in the eye region has a high impact. %, even with transparent glasses.
Indeed transparent glasses have the worst detection accuracy, even worse than shades.
Only six failed images with transparent glasses contain more than one face, the rest being due to undetected face.
One plausible reason after examination of the images is that several users wear real glasses as well (Figure~\ref{fig:detection-failures}).
This very likely creates a disturbance to the detector that is not present in images with shades, where the real glasses of the user are obfuscated with 95\% or 100\% opacity.
The synthetic nature of the glasses can also be a source of unpredictability for the detector, which has not ``seen'' this type of images beforehand.
Such AR filters do not have a smooth blending with the source image. In the case of glasses, frame pixels are set directly to zero, and it may be that the transparent glasses constitute a higher source of perturbation than the shades (consider the transition to zero both at the outer and inner sides of the frame, which is completely unnatural and unexpected in a real face image).
This is amplified even more if the user is wearing real glasses as well.
In any case,
face occlusion is a difficulty known to make detection systems to struggle \citep{https://doi.org/10.1049/bme2.12029}, with our results indicating that %although our results indicate that
eye occlusion is more critical than nose occlusion.
When the reconstruction network of Section~\ref{sect:unet} is applied to images with shades, % (Figure~\ref{fig:example_reconstruction}),
detection accuracy is recovered to a higher extent ($\sim$99\%), highlighting the benefits of the employed reconstruction.
Figure~\ref{fig:example_reconstruction} depicts the reconstruction of the two images with shades of Figure~\ref{fig:filters}, showing a clear reconstruction of the majority of the eye area in the case of shades with 95\% opacity. In the case of 100\% opacity, the reconstruction is less successful, although sufficient to obtain a good detection accuracy.

Figure \ref{fig:t-sne_clusters} details the class separation by t-SNE \citep{Maaten08tsne} of the features extracted with ResNet34.
%
%
%of the datasets after feature extraction with ResNet34  \citep{Maaten08tsne}. % (perplexity 30).
%
Only the five most frequent classes are colored due to colors' limitation.
%
%The t-SNE parameters used are the defaults of the \say{scikit-learn} library function \citep{noauthor_sklearnmanifoldtsne_nodate}. Perplexity parameter is $30$.
%
For the benchmark (unmodified) and Instagram datasets, the clusters appear well separated, which suggests that class (identity) separation is possible. % with a high accuracy.
Clusters of the dog nose dataset are still separated, although closer and with higher intra-class variability. In the datasets with glasses, and specially with shades, the clusters appear much closer. %, suggesting that identity separability in these cases will be more difficult.
It can be seen a parallelism between the t-SNE plots and the detection results of Table~\ref{tab:datasets}, in the sense that faces where the nose or eye appear obfuscated are more difficult to detect and to recognize.

We now report closed-set identification experiments in Table~\ref{tab:identification}.
%Table~\ref{tab:clf_dist_table} (distance measures) and Tables~\ref{tab:LinearSVMPerformanceValues}, ~\ref{tab:XGBoostPerformanceValues} (trained methods).
%
With distance measures, the first original (unfiltered) image is used for enrolment, and identification attempts are done with filtered images. % of the different datasets.
Comparatively, the Euclidean distance performs best, although just 1-2\% better than the other metrics. The performance on the non-filtered benchmark and Instagram datasets are the highest, with minuscule differences between them ($\approx$92-93\% for all distance measures). The dog dataset follows at $\approx$56\%, and transparent glasses at $\approx$50\%.
The performance of shades\_leak and shades\_no\_leak is poor, specially the latter one. After reconstruction, the shades\_recon\_leak dataset shows some performance recovery (66.3\%). On the other hand, reconstruction with the non-leaking shades (100\% opacity) does not contribute to any performance improvement. % with distance measures.

To carry out closed-set identification experiments with trained methods, the datasets are split into 80\% (training) and 20\% (test). Training is done either with benchmark unfiltered images (``Train=Benchmark''), with filtered images of the corresponding test dataset (``Train=Filter''), or with images of all datasets together (``Train=All'').
Identification tests are always done with filtered images.
The splits between different datasets are the same to ensure comparability, achieved by use of scikit-learn function \textit{train$\_$test$\_$split} (the hyper-parameter \textit{random$\_$state} is initialized with the same value for each split).
A first observation %from Tables~\ref{tab:LinearSVMPerformanceValues}, ~\ref{tab:XGBoostPerformanceValues}
is that both trained methods behave similarly on the different datasets, at least in those that do not involve shades. % (eye obfuscation).
%
%For the datasets
With shades, SVM is comparatively better than XGBoost, at least when ``Train=Filter'' is applied.
Also, both trained methods are widely benefited by ``Train=Filter'', % training with filtered images,
and specially by ``Train=All'', %if they are trained with all datasets together,
which provides the best results overall.
When training is done with all images, the performance of datasets not involving eye obfuscation reaches 98-100\%, and even those involving shades or glasses are recovered to 94-97\% in most cases.
Indeed, shades\_leak and shades\_no\_leak are boosted to the point of not needing reconstruction of the image before recognition.
Combining the datasets have a beneficial effect, since the classifier is 'seeing´ images of the same individual with different perturbations, acting as a way of data augmentation.
It should be considered though that %in such cases: $i$) the training dataset do contain reconstructed images, and $ii$)
the option ``Train=All'' has eight times more training data than the other options due to dataset pooling.

\begin{figure}[t]
    \centering
    \includegraphics[width=0.44\textwidth]{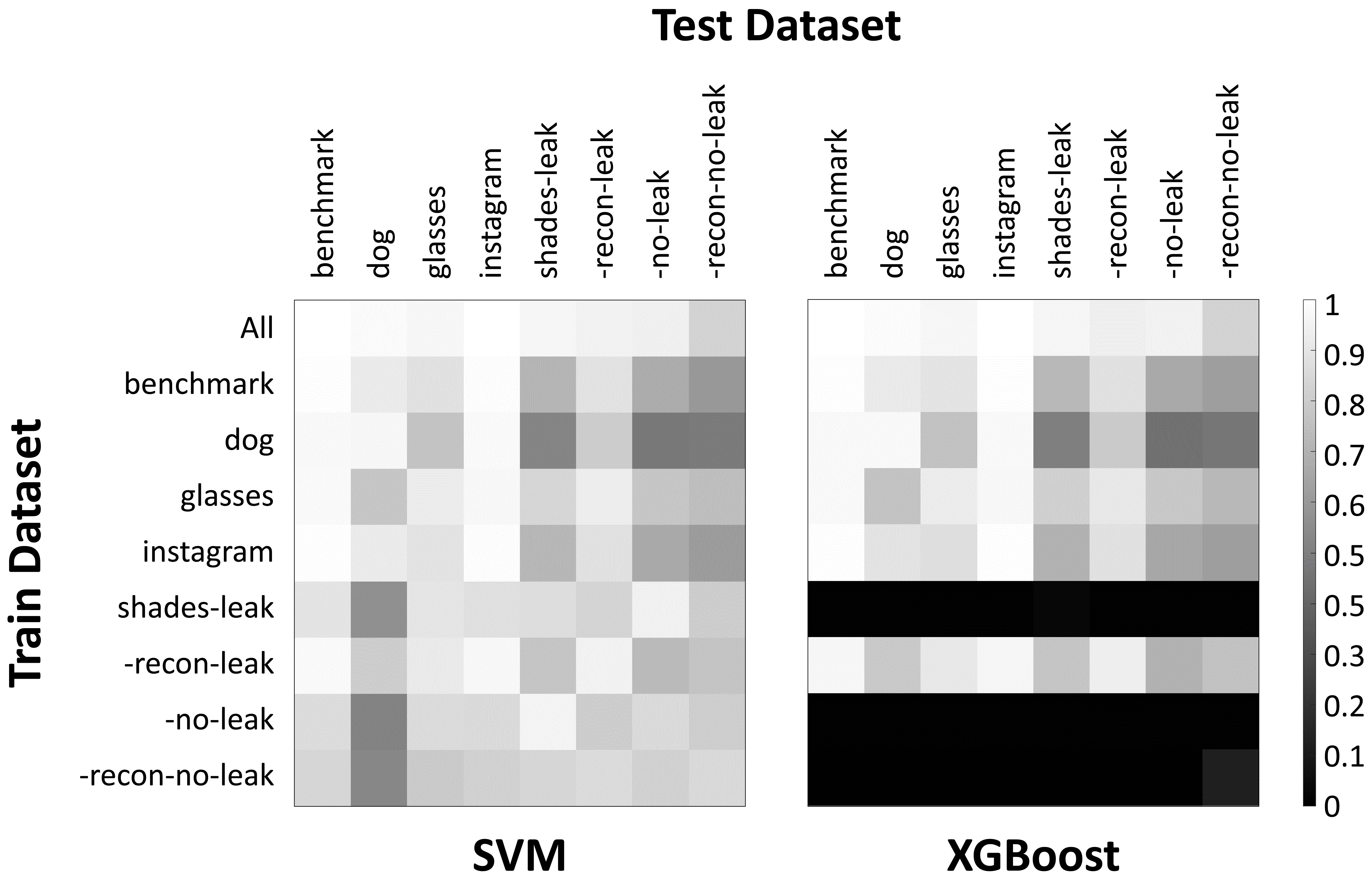}
    \caption{ResNet34: Identification Accuracy (closed set), cross-filter experiments. The higher the value, the better. The exact values are given in Table~\ref{fig:identification-crossfilter}.}
    \label{fig:identification-crossfilter}
\end{figure}

We also report in Table~\ref{tab:identification-crossfilter} cross-dataset experiments when the classifier is trained with one dataset (``Train=Filter'') and tested with another one.
For better viewing, Figure~\ref{fig:identification-crossfilter} depicts the accuracy values (black=0\%, white=100\%), including the case ``Train=All'' in the first row for comparative purposes.
It can be seen that when training does not include eye obfuscation (rows 2-5), testing with shades decreases performance significantly. Some performance recovery is achieved when testing with shades\_recon\_leak, since this reconstruction is observed to recover the eye region to a certain extent (Figure~\ref{fig:example_reconstruction}).
Another phenomenon is the poor performance of XGBoost when training involves shades (black squares). Again, the only exception is shades\_recon\_leak, but in the other three cases, this classifier is unusable.
Another interesting effect is that training with the dog dataset and testing with any glasses or shades dataset (or the opposite) produces the worst performance, apart from the XGBoost issue just mentioned (see the dark squares in row 3/column 2 of Figure~\ref{tab:identification-crossfilter}).
Dog and glasses/shades images are the most different images, in the sense that they have obfuscated different regions, so a significant portion of the face is different between training and test images. This makes that cross-filter classification struggles in identifying individuals.
In such situation, it would be problematic if the wrong classifier is used.
One way to cope with this effect, as we have seen above, is to train the classifiers with images from all datasets.

\begin{figure}[t]
    \centering
    \includegraphics[width=0.4\textwidth]{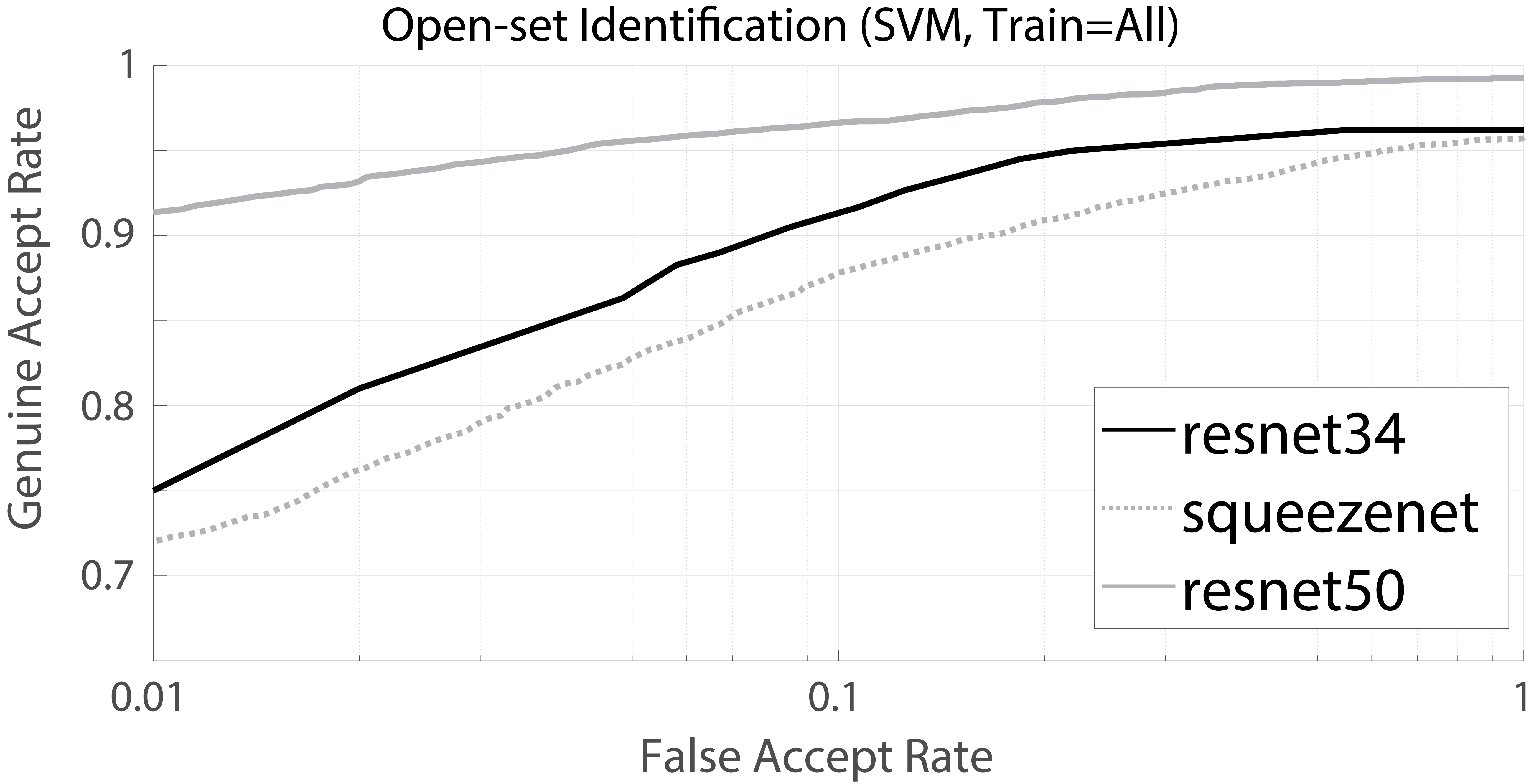}
    \caption{Identification Accuracy (open set).}
    \label{fig:identification-openset}
\end{figure}

Next, we report open-set identification experiments in Figure~\ref{fig:identification-openset}.
%
%We show Genuine Accept Rate, GAR=1-FNIR (proportion of mated searches where the enrolled template is in the top rank and the score exceeds the threshold).
%
To do these experiments, we have set aside 58 random individuals as unseen people, and trained an SVM classifier with the remaining 100 individuals.
The comparison score is the confidence measure given by the SVM.
As before, the 100 individuals are split into 80\% (training) and 20\% (test).
%
%Tests with unseen people are done using all their available images.
%
Training is done with images of all datasets together (``Train=All''), since it is the best performing option in the closed-set setting.
The amount of mated searches where the enrolled mate is in the first position of the rank is 96.2\% with ResNet34 (black curve, corresponding to the right side of the x-axis of Figure~\ref{fig:identification-openset} where the threshold is sufficiently high to allow all mated searches to exceed it; other CNN backbones will be commented later).
As we move left in the x-axis, the GAR reduces at the same time that the FAR reduces too. At FAR=10/1\%, the obtained GAR is approximately 92/75\% with ResNet34.

\begin{figure*}[htb]
    \centering
    \includegraphics[width=0.85\textwidth]{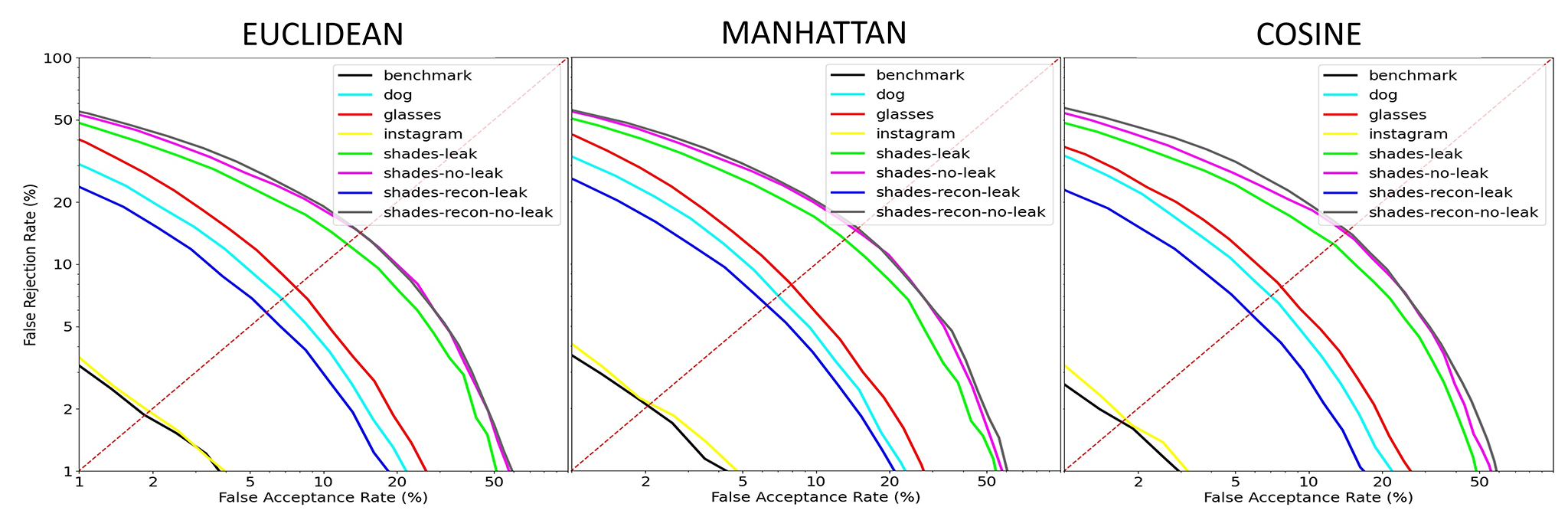}
    \caption{ResNet34: Verification results (DET curves) using Euclidean, Manhattan and Cosine distance}
    \label{fig:DET_euc}
\end{figure*}

Finally, we report verification experiments (Table~\ref{tab:EER} and Figure~\ref{fig:DET_euc}) with the different distance measures.
%
%As before,
The first original (unfiltered) image is used for enrolment, and the remaining filtered images of the different datasets for verification attempts, both genuine (mated) and impostor (non-mated).
As can be observed, the Euclidean distance performs best, although the other distances are less than 1\% behind.
%the difference w.r.t. the other distances is small (less than 1\%).
%
%The results of Table~\ref{tab:EER} shows that
The best EER is for benchmark and Instagram sets at $\approx$ 2\%, with the rest (in descending order) at 7\% (dog), 8.2\% (glasses), 12.5\% (shades\_leak), and 14\% (shades\_no\_leak).
The EER for reconstructed shades\_leak surpasses the dog results at 6\%.
Also, as before, reconstruction with the non-leaking shades (shades\_recon\_no\_leak) does not show any improvement.
%
%The complete behaviour of the datasets over the entire range of FAR and FRR values can be also see in the DET curves of Figure~\ref{fig:DET_euc}.
%
The DET curves show a similar behaviour over the entire range of FAR and FRR values, with the relative performance of the systems becoming somehow closer to each other for low FAR values.

To conclude, we report the comparison of different CNN backbones under identification and verification (Figure~\ref{fig:identification-openset} and Table~\ref{tab:cnn-comparison}).
Experiments are carried out with the best options identified previously (identification: SVM classifier, ``Train=All''; verification: cosine distance).
One immediate observation is that the much deeper ResNet50 model surpasses the results obtained previously, being the best performing backbone.
The closed-set identification accuracy is pushed towards higher values, including cases involving shades, which surpass 97\% even without reconstruction (although both reconstructed and non-reconstructed images are present in the training set).
The same can be said about verification, where the EER without any eye obfuscation is 1.6\% (dog) or 0.7\% (benchmark, instagram), and less than 3\% with transparent glasses. Shades reconstruction also provides good results, with EER=2.2\% (leak) and 5.3\% (no\_leak). The average EER of 2.9\% with ResNet50 contrasts with the 8.4\% obtained with ResNet34.
As a side note, ResNet50 and SqueezeNet backbones are trained using much larger face datasets (recall Section~\ref{sect:feature_extraction}), which may also contribute to a better face recognition performance.
This can be appreciated by the fact that SqueezeNet in Table~\ref{tab:cnn-comparison} has an identification accuracy that is just behind ResNet34, despite being a much lighter model. Indeed, the verification accuracy of SqueezeNet is better than ResNet34 in several cases, including those that entail eye obfuscation.
In open set idenfication (Figure~\ref{fig:identification-openset}), we can again see the very good performance of ResNet50 (GAR 97/91\% at FAR=10/1\%), with SqueezeNet situated behind ResNet34.

\begin{table}[htb]
\small
\caption{ResNet34: Verification EER. % (Euclidean, Manhattan and Cosine distances).
The lower the value, the better.}
\centering
\begin{tabular}{ c|c|c|c }
  \textbf{Test Dataset} & \textbf{Eucl} & \textbf{Manh} & \textbf{Cosine} \\
  \hline  \hline
 benchmark & \textbf{0.019} & 0.020  & 0.020\\  \hline
 dog & 0.071 & \textbf{0.070}  & 0.073 \\
 glasses & \textbf{0.082} & 0.085  & \textbf{0.082} \\
 instagram & \textbf{0.022} & 0.023  & 0.023 \\  \hline
 shades\_leak & \textbf{0.125} & 0.132  & \textbf{0.125} \\
 shades\_recon\_leak & 0.062 & 0.067  & \textbf{0.060} \\ \hline
 shades\_no\_leak & \textbf{0.140} & 0.147  & 0.144 \\
 shades\_recon\_no\_leak & \textbf{0.144} & 0.152  & \textbf{0.144}  \\ \hline  \hline
 \textit{average} & \textbf{0.083} & 0.087 & \textbf{0.083}

\end{tabular}
\label{tab:EER}
\end{table}

% Please add the following required packages to your document preamble:
% \usepackage{multirow}
% \usepackage{graphicx}
\begin{table}[htb]
\footnotesize

\caption{Comparison between different CNN backbones (R34=ResNet34, SQ=SqueezeNet, R50=ResNet50). Left: Identification Accuracy, closed set (the higher, the better). Right: Verification EER (the lower, the better).}

\begin{tabular}{c||ccc||ccc}

\multirow{2}{*}{} & \multicolumn{3}{c||}{\begin{tabular}[c]{@{}c@{}}Closed set Identification\\ (SVM, Train=All)\end{tabular}} & \multicolumn{3}{c}{\begin{tabular}[c]{@{}c@{}}Verification\\

(Cosine)\end{tabular}} \\ \hline

\textbf{Test Dataset} & \textbf{R34} & \textbf{SQ} & \textbf{R50} & \textbf{R34} & \textbf{SQ} & \textbf{R50} \\ \hline \hline

benchmark & \textbf{0.999} & 0.986 & \textbf{0.999} & 0.020 & 0.020 & \textbf{0.007} \\ \hline

dog & 0.982 & 0.966 & \textbf{0.992} & 0.073 & 0.035 & \textbf{0.016} \\

glasses & 0.967 & 0.917 & \textbf{0.985} & 0.082 & 0.085 & \textbf{0.028} \\

instagram & 0.998 & 0.985 & \textbf{0.999} & 0.023 & 0.020 & \textbf{0.007} \\ \hline

shades\_leak & 0.964 & 0.911 & \textbf{0.977} & 0.125 & 0.098 & \textbf{0.048} \\

*\_recon\_leak & 0.949 & 0.953 & \textbf{0.990} & 0.060 & 0.054 & \textbf{0.022} \\ \hline

*\_no\_leak & 0.941 & 0.902 & \textbf{0.972} & 0.144 & 0.100 & \textbf{0.050} \\

*\_recon\_no\_leak & 0.827 & 0.905 & \textbf{0.971} & 0.144 & 0.100 & \textbf{0.053} \\ \hline \hline

\textit{average} & 0.953 & 0.941 & \textbf{0.986} & 0.084 & 0.064 & \textbf{0.029}
\end{tabular}

\label{tab:cnn-comparison}
\end{table}

\section{Conclusions}
\label{sec:conclusions}

Social media platforms offer many different filters to \textit{beautify} selfie images %before they are uploaded
or to %. Augmented Reality (AR) filters that
modify them % the face image
by adding items like noses or glasses. % are also popular in social media and video-conference applications.
%
%
%Accordingly,
We are thus interested in studying the effect of such %popular selfie ''beautification''
filters on the accuracy of both face detection and recognition.
The effect of some of the employed filters have been observed to be detrimental to both tasks, %to face detection and identity recognition,
specially if the eye region is obfuscated.
Thus, we explore methods to reverse the applied manipulations.
Another strategy has been
%As a way of improving accuracy, we also study
the use of filtered images for enrolment.
In overall terms, by combining these two solutions, %use of filtered images to train recognition methods and the reconstruction of manipulated images,
we manage to counteract the effect of the majority of the studied image modifications.
The use of a deeper CNN backbone or a larger face dataset to pre-train the recognition backbones have been also seen as contributing factors. The latter becomes relevant for example under hardware limitations, since one of employed backbones \citep{[Alonso20SqueezeFacePoseNet]} is comparatively much shallower, but its performance is just behind deeper counterparts.

As future work, we are exploring to improve the reconstruction performance further and achieve more realistic results, for example using image translation methods based on adversarial training \citep{[isola16pix2pix],[Zhu17cycleGAN]}.
The performance of face detection itself under AR eye occlusion is another source of study, especially with transparent glasses.
Another direction not addressed is the detection of applied manipulations.
%
%Knowing which specific alteration or which face regions have been modified
This is necessary to use %reconstruction or recognition
algorithms trained on such modification specifically.
Here, we predict that detecting alterations at the patch level will be a fruitful avenue \citep{jain_detecting_2019}.
The latter can be be combined with the use of face detection or recognition methods based on local analysis, so if one particular region is occluded or altered, it is set to not contributing to the task. This is similar to, for example, using detectors of the periocular region that do not rely on the full-face being available \citep{[Alonso16]}.

\section*{Acknowledgments}

This work has been carried out by P. Hedman and V. Skepetzis in the context of their Master Thesis at Halmstad University (Master's Programme in Network Forensics).
Authors K. Hernandez-Diaz, J. Bigun and F. Alonso-Fernandez would like to thank the Swedish Research Council (VR) and the Swedish Innovation Agency (VINNOVA) for funding their research.

\bibliographystyle{model2-names}
\bibliography{References}

%\section*{Supplementary Material}

%Supplementary material that may be helpful in the review process should
%be prepared and provided as a separate electronic file. That file can
%then be transformed into PDF format and submitted along with the
%manuscript and graphic files to the appropriate editorial office.

\end{document}